\documentclass[A4paper]{article}
\usepackage{times}
\usepackage{graphicx}
\usepackage{latexsym}
\usepackage{algorithmic}
\usepackage{algorithm}
\usepackage{amsmath}

\DeclareMathOperator*{\argmax}{arg\!max}

\begin{document}

\title{Deep Incremental Boosting}

\author{Alan Mosca and George D. Magoulas
\vspace{.3cm}\\
Department of Computer Science and Information Systems\\
Birkbeck, University of London\\
Malet Street, London - UK
}
\maketitle

\newtheorem{conjecture}{Conjecture}
\newtheorem{intuition}{Intuition}
\newtheorem{assumption}{Assumption}
\newtheorem{definition}{Definition}

\begin{abstract}
This paper introduces Deep Incremental Boosting, a new technique derived
from AdaBoost, specifically adapted to work with Deep Learning methods,
that reduces the required training time and improves generalisation.
We draw inspiration from Transfer of Learning approaches to reduce the
start-up time to training each incremental Ensemble member.
We show a set of experiments that outlines some preliminary results
on some common Deep Learning datasets and discuss the potential
improvements Deep Incremental Boosting brings to traditional
Ensemble methods in Deep Learning.
\end{abstract}

\section{Introduction}
AdaBoost~\cite{schapire90} is considered a successful Ensemble method and is
commonly used in combination with traditional Machine Learning algorithms,
especially Boosted Decision Trees~\cite{dietterich2000experimental}.
One of the main principles behind it is the additional emphasis given
to the so-called \emph{hard to classify} examples from a training set.

Deep Neural Networks have also had great success on many visual
problems, and there are a number of benchmark datasets in this area
where the state-of-the-art results are held by some Deep Learning
algorithm~\cite{wan2013regularization,graham14a}.

Ideas from \emph{Transfer of Learning} have found applications in Deep Learning;
for example, in Convolutional Neural Networks (CNNs), when sub-features learned
early in the training process can be carried forward to a new CNN in order to
improve generalisation on a new problem of the same
domain~\cite{yosinski2014transferable}.  It has also been shown that these
Transfer of Learning methods reduce the ``warm-up'' phase of the training, where
a randomly-initialised CNN would have to re-learn basic feature selectors from
scratch.

In this paper, we explore the synergy of AdaBoost and Transfer of Learning to
accelerate this initial warm-up phase of training each new round of boosting.
The proposed method, named \emph{Deep Incremental Boosting}, exploits additional
capacity embedded into each new round of boosting, which  increases the
generalisation without adding much training time. When tested in Deep Learning
benchmarks, the new method is able to beat traditional Boosted CNNs on benchmark
datasets, in a shorter training time.

The paper is structured as follows. Section~\ref{Sec:PriorWork} presents an
overview of prior work on which the new development is based.
Section~\ref{Sec:DIB} presents the new learning algorithm.
Section~\ref{Sec:Experiments} reports the methodology of our preliminary
experimentation and the results.  Section~\ref{Sec:SOTA} provides examples where
state-of-the-art models have been used as the base classifiers for Deep
Incremental Boosting.  Lastly, Section~\ref{Sec:FinalRemarks} makes conclusions
on our experiments, and shows possible avenues for further development.

\section{Prior Work}
\label{Sec:PriorWork}
This section gives an overview of  previous work and algorithms on which our
new method is based.
\subsection{AdaBoost}
AdaBoost~\cite{schapire90} is a well-known Ensemble method, which has
a proven track record of improving performance. It is based on the principle of
training Ensemble members in ``rounds'', and at each round increasing the
importance of training examples that were misclassified in the previous round.
The final Ensemble is then aggregated using weights $\alpha_{0..N}$ calculated
during the training.
Algorithm~\ref{Alg:AdaBoostM2} shows the common
AdaBoost.M2~\cite{freundschapire96} variant. This variant is generally
considered better for multi-class problems, such as those used in our
experimentation, however the same changes we apply to AdaBoost.M2 can be applied
to any other variant of AdaBoost.

\begin{algorithm}
\caption{AdaBoost.M2}
\begin{algorithmic}
\STATE $m = |X_0|$
\STATE $D_0(i) = 1/m$ for all $i$
\STATE $t = 0$
\STATE $B = \{(i,y) : i \in \{1, \dots, m\}, y \neq y_i\}$
\WHILE{$t < T$}
	\STATE $X_t \gets$ pick from original training set $X_0$ with distribution $D_t$
    \STATE $h_t \gets$ train new classifier on $X_t$
    \STATE $\epsilon_{t} = \frac{1}{2} \sum_{(i,y) \in B}D_t(i) (1-h_t(x_i,y_i)+h_t(x_i,y))$
    \STATE $\beta_{t} = \epsilon_t / (1 - \epsilon_t)$
    \STATE $D_{t+1}(i) = \frac{D_t(i)}{Z_t} \cdot \beta^{(1/2)(1+h_t(x_i,y_i)-h_t(x_i,y))}$
    \STATE where $Z_t$ is a normalisation factor such that $D_{t+1}$ is a distribution
	\STATE $t = t + 1$
\ENDWHILE
\STATE $H(x) = \argmax_{y \in Y} \sum_{t=1}^T(log \frac{1}{\beta_t}h_t(x,y)$
\end{algorithmic}
\label{Alg:AdaBoostM2}
\end{algorithm}

\subsection{Transfer of Learning applied to Deep Neural Networks}
Over the last few years a lot of progress has been made in the Deep Networks 
area due to their ability to represent features at various levels of
resolution. A recent study analysed how the low-layer features of Deep Networks
are transferable and can be considered general in the problem domain of image
recognition~\cite{yosinski2014transferable}.  More specifically it has been
found that, for example, the first-layer of a CNN tends to learn filters that
are either similar to Gabor filters or color blobs.  Ref~\cite{bengio2012deep}
studies Transfer of Learning in an unsupervised setting on Deep Neural Networks
and also reached a similar conclusion.

In supervised Deep Learning contexts, transfer of learning can be achieved by
setting the initial weights of some of the layers of a Deep Neural Network to
those of a previously-trained network.  Because of the findings on the
generality of the first few layers of filters, this is traditionally applied
mostly to those first few layers. The training is then continued on the new
dataset, with the benefit that the already-learned initial features provide a
much better starting position than randomly initialised weights, and as such
the generalisation power is improved and the time required to train the network
is reduced.

\section{Deep Incremental Boosting}
\label{Sec:DIB}

\subsection{Motivation}
Traditional AdaBoost methods, and related variants, re-train a new classifier
from scratch every round. While this, combined with the weighted re-sampling of
the training set, appears at first glance to be one of the elements that create
diversity in the final Ensemble, it may not be necessary to re-initialize the
Network from scratch at every round.

It has already been previously shown that weights can be transferred between
Networks, and in particular between subsets of a network, to accelerate the
initial training phase.  In the case of Convolutional Neural Networks, this
particular approach is particularly fruitful, as the lower layers (those
closest to the input) tend to consistently develop similar features.

\subsection{Applying Transfer of Learning to AdaBoost}
\begin{intuition}
Because each subsequent round of AdaBoost increases the importance given to the
errors made at the previous round, the network $h_t$ at a given round can be
repurposed at round $t+1$ to learn the newly resampled training set.
\end{intuition}
In order for this to make sense, it is necessary to formulate a few conjectures.


\begin{definition}
    Let $X_a$ be a set composed of $n$ training example vectors
    $X_a = \{\boldsymbol{x}_{a,1}, \boldsymbol{x}_{a,2},~\dots~\boldsymbol{x}_{a,n} \}$,
    with its corresponding set of correct label vectors
    $Y_a = \{\boldsymbol{y}_{a,1}, \boldsymbol{y}_{a,2},~\dots~\boldsymbol{y}_{a,n} \}$
\end{definition}

\begin{definition}
\label{Def:mostly_similar}

A training set $X_a$ is \emph{mostly similar} to another set $X_b$
if the sets of unique instances $X_a$ and $X_b$ have more common than
different elements, and the difference set is smaller than an arbitrary
significant amount $\epsilon$.

This can be expressed equivalently as:
\begin{align}
    \left\vert X_a \cap X_b \right\vert >>
    \left\vert X_a \ominus X_b \right\vert \, \\
    \left\vert X_a \cap X_b \right\vert >>
    \left\vert X_a - X_b \right\vert + \left\vert X_b - X_a \right\vert
\end{align}
or
\begin{align}
    \left\vert X_a \cup X_b \right\vert =
    \left\vert X_a \cap X_b \right\vert + \epsilon
\end{align}
Given the Jaccard Distance 
\begin{equation}
    J(A,B) = \frac{|A \cap B|}{|A \cup B|}
\end{equation}
this can be formulated as
\begin{equation}
    J(X_a,X_b) \ge 1 - \epsilon
\end{equation}
\end{definition}
\begin{conjecture}
\label{Conj:different_sets}
At a round of Boosting $t+1$, the resampled training set $X_{t+1}$ and the
previous resampled training set $X_{t}$ are \emph{mostly similar}, as in
Definition~\ref{Def:mostly_similar}:
\begin{align}
    \left\vert X_t \cap X_{t+1} \right\vert >>
    \left\vert X_t - X_{t+1} \right\vert + \left\vert X_{t+1} - X_t \right\vert
\end{align}
or
\begin{align}
    \left\vert X_t \cup X_{t+1} \right\vert =
    \left\vert X_t \cap X_{t+1} \right\vert + \epsilon
\end{align}
\end{conjecture}

If we relax $\epsilon$ to be as large as we like, In the case of Boosting, we
know this to be true because both $X_{t}$ and $X_{t+1}$ are resampled from $X_0$
with the weighting $D_{t+1}$ from the initial dataset $X_{t=0}$, so the unique
sets $X_{t}$ and $X_{t+1}$ are large resampled subsets of the initial training
set $X_{t=0}$:

\begin{align}
    X_{t} \subseteq X_{t=0} \\
    X_{t+1} \subseteq X_{t=0} \\
    \left\vert X_t \right\vert =
    \left\vert X_{t+1} \right\vert =
    \left\vert X_{t=0} \right\vert
\end{align}

\begin{definition}
    We introduce a \emph{mistake} function $E(h_a,X_b)$ which counts the number of
    mistakes by the classifier $h_a$ on dataset $X_b$:
    \begin{align}
        E(h_a,X_b,Y) = \Bigl| \{ \boldsymbol{x}_{b,i} |
            h_a(\boldsymbol{x}_{b,i}) \ne \boldsymbol{y}_i
            \forall \boldsymbol{x}_{b,i} \in X_b \} \Bigr|
    \end{align}
    where $y_i$ is the ground truth for example $i$, taken from the correct
    label set $y$.
\end{definition}

\begin{conjecture}
\label{Conj:next_set}
Given Conjecture~\ref{Conj:different_sets} and provided that the dataset $X_t$
and $X_{t+1}$ are \emph{mostly similar} as per
Definition~\ref{Def:mostly_similar}, a classifier $h_t$ that classifies $X_t$
better than randomly will still perform better than randomly on a new dataset
$X_{t+1}$.

\end{conjecture}

Given that all sets are of the same size by definition, as they are resampled
that way, we can ignore the fact that the error count on a dataset
$E(h_t,X_t)$ would need to be divided by the size of the dataset $\left\vert X_t
\right\vert$, thus simplifying the notation.

We can therefore redefine the errors made by $h_t$ on both $X_t$ and $X_{t+1}$ as:
\begin{align}
    E(h_t, X_t, Y) = E(h_t, X_t \cap X_{t+1}, Y) + E(h_t, X_t - X_{t+1}, Y) \\
    E(h_t, X_{t+1}, Y) = E(h_t, X_t \cap X_{t+1}, Y) + E(h_t, X_{t+1} - X_t, Y)
\end{align}
From Conjecture~\ref{Conj:different_sets}, the last two terms are negligible,
leaving:
\begin{align}
    E(h_t,X_t, Y) = E(h_t, X_t \cap X_{t+1}, Y) + \epsilon_t \\
    E(h_t,X_{t+1}, Y) = E(h_t, X_t \cap X_{t+1}, Y) + \epsilon_{t+1}
\end{align}
therefore $E(h_t,X_t, Y) \approx E(h_t,X_{t+1}, Y)$.

\begin{assumption}
\label{Conj:different_networks}
The weights and structure of a classifier $h_t$ that correctly classifies the
training set $X_{t}$ will not differ greatly from the classifier $h_{t+1}$ that
correctly classifies the training set $X_{t+1}$, provided that the two sets are
\emph{mostly similar}. 
\end{assumption}

\begin{conjecture}
Given Conjecture~\ref{Conj:next_set} classifier $h_t$ and its classification
output $Y_t$, it is possible to construct a derived classifier $h_{t+1}$ that
learns the corrections on the residual set $X_{t+1} - X_t$.
\end{conjecture}

When using Boosting in practice, we find these assumptions to be true most of
the time.  We can therefore establish a procedure by which we preserve the
knowledge gained from round $t$ into the next round $t+1$:

\begin{enumerate}

\item At $t = 0$, a new CNN is trained with random initialisations on the
    re-sampled dataset $X_0$, for $N$ iterations.

\item The new dataset $X_{t+1}$ is selected. The calculation of the error
    $\epsilon_t$, the sampling distribution $D_t$ and the classifier weight
    $\alpha_t$ remain the same as per AdaBoost.M2.

\item At every subsequent round, the struture of network $h_t$ is copied and
    extended by one additional hidden layer, at a given position in the network
    $i$, and all the layers below $i$ are copied into the new network.  By doing
    so, we preserve the knowledge captured in the previous round, but allow for
    additional capacity to learn the corrections on $X_{t+1} - X_t$.  This new
    network is trained for $M$ iterations, where $M << N$.  \item Steps $2$ and $3$
    are repeated iteratively until the number of rounds has been exhausted.
\end{enumerate}

Because the Network $h_{t+1}$ doesn't have to re-learn basic features, and
already incorporates some knowledge of the dataset, the gradients for the lower
layers will be smaller and the learning will be concentrated on the newly added
hidden layer, and those above it.
This also means that all classifiers $h_{t>1}$
will require a smaller number of epochs to converge, because many of the weights
in the network are already starting from a favourable position to the dataset.

At test time, the full group of hypotheses is used, each with its respective
weight $\alpha_t$, in the same way as AdaBoost.

Algoritm~\ref{Alg:DIB} shows the full algorithm in detail.

\begin{algorithm}
\caption{Deep Incremental Boosting}
\begin{algorithmic}
\STATE $D_0(i) = 1/M$ for all $i$
\STATE $t = 0$
\STATE $W_0 \gets $ randomly initialised weights for first classifier
\WHILE{$t < T$}
	\STATE $X_t \gets$ pick from original training set with distribution $D_t$
	\STATE $u_t \gets$ create untrained classifier with additional layer of shape $L_{new}$
	\STATE copy weights from $W_t$ into the bottom layers of $u_t$
    \STATE $h_t \gets$ train $u_t$ classifier on current subset
    \STATE $W_{t+1} \gets$ all weights from $h_t$
    \STATE $\epsilon_{t} = \frac{1}{2} \sum_{(i,y) \in B}D_t(i) (1-h_t(x_i,y_i)+h_t(x_i,y))$
    \STATE $\beta_{t} = \epsilon_t / (1 - \epsilon_t)$
    \STATE $D_{t+1}(i) = \frac{D_t(i)}{Z_t} \cdot \beta^{(1/2)(1+h_t(x_i,y_i)-h_t(x_i,y))}$
    \STATE where $Z_t$ is a normalisation factor such that $D_{t+1}$ is a distribution
    \STATE $\alpha_t = \frac{1}{\beta_t}$
	\STATE $t = t + 1$
\ENDWHILE
\STATE $H(x) = \argmax_{y \in Y} \sum_{t=1}^T(log \alpha_t h_t(x,y)$
\end{algorithmic}
\label{Alg:DIB}
\end{algorithm}

\section{Experimental Analysis}
\label{Sec:Experiments}
Each experiment was repeated $20$ times, both for AdaBoost.M2 and Deep
Incremental Boosting, using the same set of weight initialisations (one for each
run), so that any possible fluctuation due to favourable random starting
conditions was neutralised. Each variant ran for a fixed $10$ rounds of
boosting. We trained each Ensemble member using Adam\cite{kingma2014adam}, and
used a hold-out validation set to select the best model.

All the experiments were run on an Intel Core i5 3470 cpu with a nVidia GTX1080
GPU using the toupee Ensemble library available online at
\url{https://github.com/nitbix/toupee}.

Code and parameters for these experiments is available online at
\url{https://github.com/nitbix/ensemble-testing}.

\subsection{Datasets}

\subsubsection{MNIST}
MNIST~\cite{mnistlecun} is a common computer vision dataset that associates
pre-processed images of hand-written numerical digits with a class label
representing that digit. The input features are the raw pixel values for the $28
\times 28$ images, in grayscale, and the outputs are the numerical value between
$0$ and $9$.

The CNN used for MNIST has the following structure:

\begin{itemize}
    \item An input layer of $784$ nodes, with no dropout
    \item $64$ $5 \times 5$ convolutions, with no dropout
    \item $2 \times 2$ max-pooling
    \item $128$ $5 \times 5$ convolutions, with no dropout
    \item $2 \times 2$ max-pooling
    \item A fully connected layer of $1024$ nodes, with $50\%$ dropout
    \item a Softmax layer with $10$ outputs (one for each class)
\end{itemize}

This network has $\approx 2.3$ million weights.

The layer added during each round of Deep Incremental Boosting
is a convolutional layer of $64$ $3 \times 3$ channels, with no dropout,
added after the second max-pooling layer.

\subsubsection{CIFAR-10}
CIFAR-10 is a dataset that contains $60000$ small images of $10$ categories of
objects. It was first introduced in \cite{krizhevsky2009learning}. The images
are $32 \times 32$ pixels, in RGB format. The output categories are
\emph{airplane, automobile, bird, cat, deer, dog, frog, horse, ship, truck}. The
classes are completely mutually exclusive so that it is translatable to a
\emph{1-vs-all} multiclass classification.
Of the $60000$ samples, there is a training set of $40000$ instances, a
validation set of $10000$ and a test set of another $10000$. All sets have
perfect class balance.

The CNN used for CIFAR-10 has the following structure:

\begin{itemize}
    \item An input layer of $3096$ nodes, with no dropout
    \item $64$ $3 \times 3$ convolutions, with $25\%$ dropout
    \item $64$ $3 \times 3$ convolutions, with $25\%$ dropout
    \item $2 \times 2$ max-pooling
    \item $128$ $3 \times 3$ convolutions, with $25\%$ dropout
    \item $128$ $3 \times 3$ convolutions, with $25\%$ dropout
    \item $2 \times 2$ max-pooling
    \item $256$ $3 \times 3$ convolutions, with $25\%$ dropout
    \item $256$ $3 \times 3$ convolutions, with $25\%$ dropout
    \item $2 \times 2$ max-pooling
    \item A fully connected layer of $1024$ nodes, with $50\%$ dropout
    \item a Softmax layer with $10$ outputs (one for each class)
\end{itemize}

This network has $\approx 5.4$ million weights.

The layer added during each round of Deep Incremental Boosting
is a convolutional layer of $128$ $3 \times 3$ channels, with no dropout,
added after the second max-pooling layer.

\subsection{CIFAR-100}
CIFAR-100 is a dataset that contains $60000$ small images of $100$ categories of
objects, grouped in $20$ super-classes. It was first introduced in
\cite{krizhevsky2009learning}. The image format is the same as CIFAR-10.
Class labels are provided for the $100$ classes as well as the $20$
super-classes. A super-class is a category that includes $5$ of the
fine-grained class labels (e.g. ``insects'' contains \emph{bee, beetle,
butterfly, caterpillar, cockroach}).
Of the $60000$ samples, there is a training set of $40000$ instances, a
validation set of $10000$ and a test set of another $10000$. All sets have
perfect class balance.

The model we used has the same structure as the one we trained on CIFAR-10.

\subsection{Results}

We can see from these preliminary results in Table~\ref{Tab:Results} that
Deep Incremental Boosting is able to generalise better than AdaBoost.M2.
We have also run AdaBoost.M2 with larger CNNs, up to the size of the largest
CNN used in Deep Incremental Boosting (e.g. with 10 additional layers)
and found that the classification performance was gradually getting worse as
the weak learners were overfitting the training set.
We therefore assume that the additional capacity alone was not sufficient
to justify the improved generalisation and it was specifically due to the
transferred weights from the previous round, and the new layer learning the
``corrections'' from the new training set.

\begin{table}
\centering
\begin{tabular}{ l | l l l }
\hline

 & Single Network & AdaBoost.M2 & Deep Incremental Boosting \\
\hline
CIFAR-10  & $25.10$ \% & $23.57$ \% & $19.37$ \% \\
CIFAR-100 & $58.34$ \% & $57.09$ \% & $53.49$ \% \\
MNIST     &  $0.68$ \% &  $0.63$ \% &  $0.55$ \% \\

\hline
\end{tabular}
\caption{Mean misclassification rate on the test set}
\label{Tab:Results}
\end{table}

\subsection{Training Time}
We see from Table~\ref{Tab:BestEpoch} that with Deep Incremental Boosting the
best validation error is reached much earlier during the last boosting round.
This confirms our observation in Section~\ref{Sec:DIB} that the learning would
converge at an earlier epoch at subsequent rounds ($t > 1$).
Based on this we have used a shorter training schedule for these subsequent
rounds, which means that we were able to save considerable time compared to the
original AdaBoost, even though we have trained a network with a larger number of
parameters.  A summary of the improved training times is provided in
Table~\ref{Tab:TrainingTimes}.

\begin{table}
\centering
\begin{tabular}{ l | l l }
\hline
& AdaBoost.M2 & Deep Incremental Boosting \\
\hline
CIFAR-100 & $19$ & $6$ \\
CIFAR-10  & $18$ & $4$ \\
MNIST    & $14$ & $3$ \\
\hline
\end{tabular}
\caption{Typical ``best epoch'' during the $10^{th}$ round of Boosting}
\label{Tab:BestEpoch}
\end{table}

\begin{table}
\centering
\begin{tabular}{ l | l l }
\hline
& AdaBoost.M2 & Deep Incremental Boosting \\
\hline
CIFAR-100 & $\approx 26$hrs & $\approx 8$hrs \\
CIFAR-10  & $\approx  9$hrs & $\approx 3$hrs \\
MNIST    & $\approx  4$hrs & $\approx 1$hrs \\
\hline
\end{tabular}
\caption{Mean training times for each dataset}
\label{Tab:TrainingTimes}
\end{table}

\section{Larger models}
\label{Sec:SOTA}

The base classifiers we used in the experimentation in
Section~\ref{Sec:Experiments} are convenient for large numbers of repetitions
with lock-stepped random initialisations, because they train relatively quickly.
The longest base classifier to train is the one used for CIFAR-100 and it took
$\approx 3$ hours. However, these models give results that are still far from
the state-of-the-art, so we experimented further with some of these more
complicated models and applied Deep Incremental Boosting.

Because of the time required to train each model, and the differences in the
problem setup, we have not been able to run them with the same schedule as the
main experiments, therefore they have been documented separately.

\subsection{MNIST}
The best result on MNIST that doesn't involve data augmentation or manipulation
is obtained by applying Network in Network~\cite{lin2013network}. In the paper,
a full model is described, which we have been able to reproduce. Because our
goal is to train Ensembles quickly, we reduced the training schedule to $100$
epochs and applied Adam as the update rule, which also sped up the training
significantly. This network has a total of $\approx 0.35$ million weights,
however, it has a significantly higher number of computations.

After the first dropout layer, we added a new convolutional layer of $64$ $5
\times 5$ filters, at each Deep Incremental Boosting round.

\begin{table}
\centering
\begin{tabular}{ l | l l }
\hline
Method & Mean Test Misclassification & Mean Training Time \\
\hline
NiN & $0.46$\% & $\approx 18$ min \\
AdaBoost.M2  & $0.47$\% & $\approx 207$ min  \\
DIB     & $0.42$\%  & $\approx 38$ min\\
\hline
\end{tabular}
\caption{Network-in-Network results on MNIST}
\label{Tab:MNISTsota}
\end{table}

Table~\ref{Tab:MNISTsota} shows that, although the remaining examples to be
learned are very few, DIB is able to improve where AdaBoost no longer offers any
benefits. In addition to this, the training time has been reduced significantly
compared to AdaBoost.

\subsection{CIFAR-10}
The published models that achieve state-of-the-art performance on CIFAR-10 and
CIFAR-100 do not make use of a hold-out validation set. Instead, they use the
additional $10000$ examples as additional training data. In order to reproduce
similar test error results, the same principle was applied to this experimental
run.

A very efficient model of \emph{all- convolutional networks} has been proposed,
with state-of-the-art results on the CIFAR-10 dataset, which replaces the
max-pooling with an additional convolution with stride $s > 1$, and does not use
a fully-connected layer after the convolutions~\cite{springenberg2014striving}.
Instead, there are further convolutions to reduce the dimensionality of the
output, until it is possible to perform Global Average Pooling.
We based our larger model on this architecture, but in order to make the
computations feasible for an Ensemble we had to modify it slightly. The final
structure of the network is as follows:

\begin{itemize}
    \item An input layer of $3096$ nodes, with no dropout
    \item $128$ $3 \times 3$ convolutions, with $25\%$ dropout
    \item $128$ $3 \times 3$ convolutions, with $25\%$ dropout
    \item $128$ $3 \times 3$ convolutions, with a stride length of $2$
    \item $256$ $3 \times 3$ convolutions, with $25\%$ dropout
    \item $256$ $3 \times 3$ convolutions, with $25\%$ dropout
    \item $256$ $3 \times 3$ convolutions, with a stride length of $2$
    \item $512$ $3 \times 3$ convolutions, with $25\%$ dropout
    \item $512$ $3 \times 3$ convolutions, with $25\%$ dropout
    \item $512$ $3 \times 3$ convolutions, with a stride length of $2$
    \item $2 \times 2$ max-pooling
    \item A fully connected layer of $1024$ nodes, with $50\%$ dropout
    \item a Softmax layer with $10$ outputs (one for each class)
\end{itemize}
This network has $\approx 9.4$ million weights and is considerably harder to
train than the one in the original experiment. The results are reported in
Table~\ref{Tab:CIFAR-10sota}, including training time and a comparison with
vanilla AdaBoost.

\begin{table}
\centering
\begin{tabular}{ l | l l }
\hline
Method & Mean Test Misclassification & Mean Training Time \\
\hline
Single Network & $16.10$\% & $104$ mins \\
AdaBoost.M2    & $16.00$\% & $320$ mins \\
DIB            & $15.10$\% & $220$ mins \\
\hline
\end{tabular}
\caption{All-CNN results on CIFAR-10}
\label{Tab:CIFAR-10sota}
\end{table}

Each original member was trained for $40$ epochs, while each round of Deep
Incremental Boosting after the first was only trained for $20$ epochs. No
additional layer was created, due to GPU memory limitations, which is why the
improvement is not as dramatic as seen in the original experiments. However, the
time improvement alone is sufficient to justify using this new method.

\section{Concluding Remarks}
\label{Sec:FinalRemarks}

In this paper we have introduced a new algorithm, called Deep Incremental
Boosting, which combines the power of AdaBoost, Deep Neural Networks and
Transfer of Learning principles, in a Boosting variant which is able to improve
generalisation.  We then tested this new algorithm and compared it to
AdaBoost.M2 with Deep Neural Networks and found that it generalises better on
some benchmark image datasets, further supporting our claims.

One final observation that can be made is about the fact that we are still using
the entire Ensemble at test time.  In certain situations, it has been shown that
a small model can be trained to replicate a bigger one, without significant loss
of generalisation~\cite{ba2013deep}.  In future work we will investigate the
possibility to modify Deep Incremental Boosting such that only one final
test-time Deep Neural Network will be necessary.

\bibliographystyle{plain}
\bibliography{biblio}

\begin{thebibliography}{10}

\bibitem{ba2013deep}
Lei~Jimmy Ba and Rich Caurana.
\newblock Do deep nets really need to be deep?
\newblock {\em Advances in neural information processing systems}, pages
  2654--2662, 2014.

\bibitem{bengio2012deep}
Yoshua Bengio.
\newblock Deep learning of representations for unsupervised and transfer
  learning.
\newblock {\em Unsupervised and Transfer Learning Challenges in Machine
  Learning}, 7:19, 2012.

\bibitem{dietterich2000experimental}
Thomas~G Dietterich.
\newblock An experimental comparison of three methods for constructing
  ensembles of decision trees: Bagging, boosting, and randomization.
\newblock {\em Machine learning}, 40(2):139--157, 2000.

\bibitem{graham14a}
Benjamin Graham.
\newblock Fractional max-pooling.
\newblock {\em CoRR}, abs/1412.6071, 2014.

\bibitem{kingma2014adam}
Diederik Kingma and Jimmy Ba.
\newblock Adam: A method for stochastic optimization.
\newblock {\em arXiv preprint arXiv:1412.6980}, 2014.

\bibitem{krizhevsky2009learning}
Alex Krizhevsky and Geoffrey Hinton.
\newblock Learning multiple layers of features from tiny images, 2009.

\bibitem{mnistlecun}
Yann Lecun and Corinna Cortes.
\newblock {The MNIST database of handwritten digits}.

\bibitem{lin2013network}
Min Lin, Qiang Chen, and Shuicheng Yan.
\newblock Network in network.
\newblock {\em arXiv preprint arXiv:1312.4400}, 2013.

\bibitem{schapire90}
R.~E. Schapire.
\newblock The strength of weak learnability.
\newblock {\em Machine Learning}, 5:197--227, 1990.

\bibitem{freundschapire96}
R.~E. Schapire and Y~Freund.
\newblock Experiments with a new boosting algorithm.
\newblock {\em Machine Learning: proceedings of the Thirteenth International
  Conference}, pages 148--156, 1996.

\bibitem{springenberg2014striving}
Jost~Tobias Springenberg, Alexey Dosovitskiy, Thomas Brox, and Martin
  Riedmiller.
\newblock Striving for simplicity: The all convolutional net.
\newblock {\em arXiv preprint arXiv:1412.6806}, 2014.

\bibitem{wan2013regularization}
Li~Wan, Matthew Zeiler, Sixin Zhang, Yann~L Cun, and Rob Fergus.
\newblock Regularization of neural networks using dropconnect.
\newblock In {\em Proceedings of the 30th International Conference on Machine
  Learning (ICML-13)}, pages 1058--1066, 2013.

\bibitem{yosinski2014transferable}
Jason Yosinski, Jeff Clune, Yoshua Bengio, and Hod Lipson.
\newblock How transferable are features in deep neural networks?
\newblock In {\em Advances in Neural Information Processing Systems}, pages
  3320--3328, 2014.

\end{thebibliography}
\end{document}